\documentclass[11pt,a4paper]{article}          
\usepackage[dvips]{graphicx}
\usepackage{graphics}

\usepackage[affil-it]{authblk}

\usepackage{quoting}

\usepackage[T1]{fontenc}
\usepackage{lmodern}
\usepackage[utf8]{inputenc}
\usepackage[british]{babel}
\usepackage[cmex10]{amsmath}
\usepackage{amsfonts, amssymb}
\usepackage[amsmath,thmmarks]{ntheorem}
\usepackage[mathcal]{euscript}
\usepackage{bm}

\usepackage[ruled,linesnumbered]{algorithm2e}

\usepackage{caption}
\usepackage[table]{xcolor}
\usepackage{array}

\usepackage{enumitem}
\usepackage[hyperfootnotes=false,colorlinks=false,hidelinks]{hyperref}
\usepackage{url}
\usepackage{acronym}

\usepackage{flushend}
\usepackage{color,soul}
\definecolor{lightblue}{rgb}{0.9,0.95,1}
\definecolor{orange}{rgb}{1,.65,0}

\soulregister\cite7

\newcommand{\x}{\bm{x}}

\newcommand{\w}{\bm{w}}
\renewcommand{\d}{\text{d}}

\newcommand{\SE}{\mathbb{E}}

\newcommand{\SH}{\mathcal{H}}
\newcommand{\SP}{\mathcal{P}}
\newcommand{\SX}{\mathcal{X}}

\newcommand{\SZ}{\mathcal{Z}}

\usepackage{dsfont}
\newcommand{\1}{\mathds{1}}

\newcommand{\Z}{\mathbb{Z}}

\newcommand{\R}{\mathbb{R}}
\newcommand{\X}{\bm{X}}
\newcommand{\Y}{\bm{Y}}

\mathchardef\breakingcomma\mathcode`\,
{\catcode`,=\active
  \gdef,{\breakingcomma\discretionary{}{}{}}
}
\newcommand{\mathlist}[1]{$\{\mathcode`\,=\string"8000 #1\}$}
\newcommand{\etal}{\emph{et al.} }

\newcommand{\meqref}[1]{Eq.~\eqref{#1}}
\newcommand{\fref}[1]{Figure~\ref{#1}}

\newcommand{\tref}[1]{Table~\ref{#1}}
\newcommand{\sref}[1]{Section~\ref{#1}}

\newcommand{\remp}[1]{\widehat{R}_{#1}}

\theoremstyle{plain}
\theoremheaderfont{\normalfont\bfseries}\theorembodyfont{\slshape}
\theoremseparator{:}

\newtheorem{Def}{Definition}

\newtheorem{Example}{Example}

\newcommand\xqed[1]{%
  \leavevmode\unskip\penalty9999 \hbox{}\nobreak\hfill
  \quad\hbox{#1}}
\newcommand\demo{\xqed{$\triangle$}}

\begin{document}

\title{Distribution-Based Categorization of Classifier Transfer Learning}


\author{Ricardo~Gamelas~Sousa}
\affil{Farfetch, Porto, Portugal \\
\texttt{rsousa@rsousa.org}
}

\author{Lu\'{i}s~A.~Alexandre}
\affil{Instituto de Telecomunica\c{c}\~{o}es, Universidade da Beira Interior, Covilh\~{a}, Portugal \\
\texttt{lfbaa@ubi.pt} 
}

\author{Jorge~M.~Santos}
\affil{Instituto Superior de Engenharia, Polit\'{e}cnico do Porto, Portugal \\
\texttt{jms@isep.ipp.pt}
}

\author{Lu\'{i}s~M.~Silva}
\affil{Dep. de Matem\'{a}tica at Universidade de Aveiro, Portugal \\
\texttt{lmas@ua.pt}
}

\author{Joaquim~Marques~de~S\'{a}}
\affil{Instituto de Engenharia Biom\'{e}dica (INEB), Porto, Portugal}

\acrodef{CNN}{Convolutional Neural Network}
\acrodef{DA}{Domain Adaption}
\acrodef{DL}{Direct Learning}
\acrodef{DNN}{Deep Neural Network}
\acrodefplural{DNN}[DNNs]{Deep Neural Networks}
\acrodef{KL}{Kullback-Leibler}
\acrodef{knn}{k-Nearest Neighbors}
\acrodef{JS}{Jensen-Shannon}
\acrodef{ML}{Machine Learning}
\acrodef{MTL}{Multi-Task Learning}
\acrodef{NN}{Neural Network}
\acrodefplural{NN}[NNs]{Neural Networks}
\acrodef{NLP}{Natural Language Processing}
\acrodef{PoS}{Part of Speech}
\acrodef{r.v.}{random variable}
\acrodef{SVM}{Support Vector Machine}
\acrodef{SDA}{Stacked Denoising Autoencoder}
\acrodef{TL}{Transfer Learning}

\maketitle

\begin{abstract}
\ac{TL} aims to transfer knowledge acquired in one problem, the source problem, onto another problem, the target problem, dispensing with the bottom-up construction of the target model. Due to its relevance, \ac{TL} has gained significant interest in the \ac{ML} community since it paves the way to devise intelligent learning models that can easily be tailored to many different applications. As it is natural in a fast evolving area, a wide variety of \ac{TL} methods, settings and nomenclature have been proposed so far. However, a wide range of works have been reporting different names for the same concepts. This concept and terminology mixture contribute however to obscure the \ac{TL} field, hindering its proper consideration. In this paper we present a review of the literature on the majority of classification \ac{TL} methods, and also a distribution-based categorization of \ac{TL} with a common nomenclature suitable to classification problems. Under this perspective three main TL categories are presented, discussed and illustrated with examples.
\end{abstract}


\section{Introduction}\label{intro}
One common difficulty arising in practical machine learning applications is the need to redesign the classifying machines (classifiers) whenever the respective probability distributions of inputs and outputs change, even though they may relate to similar problems.
For instance, classifiers that perform recommendations of consumer items for the Amazon website cannot be straightforwardly applied to the IMDB website~\cite{PereiraML2010,Bruzzone2010}.
In the same way, text classification for the Wikipedia website may not perform appropriately when applied to the Reuters website, even though the texts of one and the other are written in the same idiom with only moderate changes of the text statistics.
The \emph{reuse} of a classifier designed for a given (\emph{source}) problem on another (\emph{target}) problem, presenting some similarities with the original one, with only minor operations of parameter tuning, is the scope of \ac{TL}.

The following aspects have recently contributed to the emergence of \ac{TL}:
\begin{itemize}
\item Considerable amount of unlabeled data: \ac{TL} relaxes the necessity of obtaining large amounts of labeled data for new problems~\cite{PereiraML2010,Telmo2014b}. \ac{TL} can be advantageous since unlabeled data can have severe implications in some fields of research, such as in the biomedical field~\cite{marx2013,Yang2013};
\item Good generalization: \ac{TL} often produces algorithms with good generalization capability for different problems~\cite{PereiraML2010,Orabona2013};
\item Less computational effort: \ac{TL} provides learning models (classifier models) that can be applied with good performance results in different problems and far less computational effort (see e.g.,~\cite{Chetak2014a,Chetak2014b,PereiraML2010}).
\end{itemize}
The following definitions clarify what we believe should be meant by \ac{TL}. We use ``model'' as a general designation of a classifier or regressor, although in the present paper we restrict ourselves to the reusability of \emph{classifiers}.
\begin{Def}[MK]
\label{def:MK}
Model Knowledge, or simply \emph{knowledge} when no confusion arises, means the functional form of a model and/or a subset of its parameters.
\end{Def}
\begin{Def}[TL]
\label{def:TL}
TL is a \ac{ML} research field whose goal is the development of algorithms capable of transferring knowledge from the source problem model in order to build a model for a target problem.
\end{Def}
Although (portions of) these ideas have been around in the literature~\cite{Yang2013}, it has never been clearly defined as Definition \ref{def:TL}.
For more than three decades there has been a significant amount of work on \ac{TL} but without a clear definition of what \ac{TL} is in general.
Most of the times the concept of \ac{TL} has been mixed in the literature with active, online~\cite{Bishop2007} and even sequential learning~\cite{Bishop2007,Quionero2009}; also, concepts from classical statistical learning theory have not been used to properly define all possible \ac{TL} scenarios. \ac{TL} in fact encompasses ideas from areas such as \emph{dataset shift}~\cite{Quionero2009} where the distribution of the data can change over time, and to which sequential algorithms may be applied; or \emph{covariate shift}~\cite{PereiraML2010} where data distributions of two problems differ but share the same decision functions.
Overall, the aforementioned discrepancies (e.g., terminology mixture) contribute to obscure the \ac{TL} field while hindering its proper consideration.

The paper is structured as follows:
\sref{sec:twotwo} presents an historical overview on the developments of \ac{TL};
\sref{sec:two} describes the fundamental concepts of \ac{TL} and their derivations followed by our analysis of \ac{TL} with illustrative examples;
In \sref{sec:issues} we list the most recent open issues on \ac{TL};
\sref{sec:five} concludes this manuscript by summing up and discussing all main ideas.

\section{Previous Work on Transfer Learning}\label{sec:twotwo}
\ac{TL} has been around since the 80's with considerable advancements since then (see e.g.,~\cite{mitchell1980,Vapnik1999,BScholkopf2006,PereiraML2010,Bengio2011,Tommasi_PAMI2013,caputo2014} and references therein). Probably, the first work that envisioned the concept of \ac{TL} was the one of Mitchel in~\cite{mitchell1980} where the idea of bias learning was presented. The first ideas for what is now known as \ac{TL} were drawn in this work.

An early attempt to extend these ideas was soon performed by Pratt \etal in~\cite{Pratt1992} where \acp{NN} were used for \ac{TL}.
In a simple way, layer weights trained for the source problem (also coined as \emph{in-domain}) were reused and retrained to solve the target problem (also coined as \emph{out-domain}).\footnote{We stick to the source and target problem.}
At the time, Pratt and her collaborators~\cite{Pratt1992} adopted entropy measures to assess the quality of the hyperplanes tailored for the target problem and to define stopping criteria for training \acp{NN} for \ac{TL}. Soon after, Intrator~\cite{intratorcs1996} derived a framework to use (abstract) internal representations generated by \acp{NN} on the source problem to solve the target problem.

After these pioneer works, a significant number of implementations and derivations of \ac{TL} started to appear.
In~\cite{Thrun96a} a new learning paradigm was proposed for \ac{TL} where one would incrementally learn concept after concept. Thrun~\cite{Thrun96a} envisioned this approach to how humans learn: by stacking knowledge upon another (as building blocks) resulting in an extreme nested system of learning functions.
At that time, a particular case of~\cite{Thrun96a}, coined \ac{MTL}, was presented~\cite{Caruana1997,BaxterJAIR2000}.
In short, \ac{MTL} solves all target problems at once using a single learning model.
To address \ac{MTL} usual \acp{NN} are employed with variations in their architectures.
Input layers are usually maintained as in standard approaches of \acp{NN}, but hidden layers can be defined as a pyramidal structure and output layers are given for each target problem. Intuitively, a \ac{NN} for \ac{MTL} would learn common representations for all problems but the decision would be addressed distinctly by each output layer~\cite{Caruana1997}.
However, this approach does not hold for our definition of \ac{TL} (see Definition \ref{def:TL}) since it learns a common representation for all problems (multiple target problems addressed simultaneously).

The concept of \emph{covariate shift} was introduced by Shimodaira~\cite{shimodaira2000}. Although initially not contextualized in the domain of \ac{TL}, his theoretical conclusions on how to learn a regression model on a target problem based on a source problem had a significant impact later to be realized.
Shimodaira described a weighted least squares regressor based on the prior knowledge of the densities of source and target problems. At the time, he only addressed the case of marginal distributions being different and equal posterior leading to what he termed \emph{covariate shift}. 
Other authors~\cite{BScholkopf2006,Marcu2006,Pereira2007} followed with different algorithms to address the limitations of Shimodaira's~\cite{shimodaira2000} work such as the estimation of data densities leading to the rise of the \emph{domain adaptation}.\footnote{Domain adaptation has similar principles to \emph{covariate shift}. We stick to the latter designation.} In the work of Sugiyama \etal~\cite{sugiyama2007} an extension of Shimodaira's work was presented so that it could cope with the leave-one-out risk.
The success of \emph{covariate shift} is mostly associated to solving many \ac{NLP} problems~\cite{Pereira2007,Pereira2007amazon,duan2012domain,liu2014}. In fact, many other works were dedicated to this subject---see e.g.,~\cite{bacchiani2003,scholkopf2006,Blitzer2006,Ling2008,gretton2009covariate,ZhangSIGKDD2009,scholkopf2009,Bruzzone2010} and references therein.
Recently, an overview on \ac{TL} was presented by Pan \etal in~\cite{Pan2010} with a vast but horizontal analysis of the most recent works that tackle classification, regression and unsupervised learning for \ac{TL}.
Orabona and co-workers\cite{Orabona2013} provided fundamental mathematical reasonings for \ac{TL} by devising: 1) generalization bounds for max-margin algorithms such as SVMs and 2) their theoretical bounds based on the leave-one-out risk~\cite{bousquet2002}. This was afterwards extended by Ben-David \etal in~\cite{BenDavid2013}.
The work of Orabona \etal~\cite{Orabona2013} was the first to identify a gap in the literature of the theoretical limitations of algorithms on \ac{TL}.

The majority of the aforementioned works assume equal number of classes both for source and target problems.
A significant contribution to the unconstrained scenario of the class set was presented in~\cite{Tommasi_PAMI2013} expanding the work of Thrun~\cite{Thrun96a}.

With the recent re-interest on \acp{NN} and the availability of more computational power along with new and faster algorithms, \acp{NN} with deep architectures started to emerge to tackle \ac{TL}.
In~\cite{Bengio2011} a framework for \emph{covariate shift} with deep networks was presented; In~\cite{Chetak2014a,Chetak2014b,Telmo2014a,Telmo2014b} the research line of~\cite{Pratt1992} was widened by addressing the following questions: How can one tailor \acp{DNN} for \ac{TL}? How does \ac{TL} perform by reusing layers and using different types of data?

The immense diversity of \ac{TL} interpretations and definitions gave rise to concerns on how to unify this area of research.
To this respect, Patricia \etal proposed in~\cite{caputo2014} an algorithm to solve \emph{covariate shift} and other types of \ac{TL} settings. In what follows we present a theoretical framework that ties together most of the work presented so far on \ac{TL} for classification problems.

\section{Classifier Transfer Learning}\label{sec:two}

%

\subsection{Classification: Notation and Problem Setting}\label{sec:notations}

A dataset represented by a set of tuples $D=\{(\x_i,y_i)\}_{i=1}^N$ is given to a classifier learning machine.
The set $\X_N = \{\x_1,..., \x_N\}$ contains $N$ instances (realizations) of a random vector $\X$ whose codomain is $\X = \R^d$; it will be clear from the context if $\X$ denotes a codomain or a random vector. Any instance $\x \in \X$ is a $d$-dimensional vector of real values $\x = [x_1, x_2, \ldots,x_j,\ldots,x_d]^t$. Similarly, the set $\Y_N = \{y_1,\ldots,y_N\}$ contains $N$ instances of a one-dimensional random variable $\Y$ whose codomain is w.l.o.g. $\Y=\Z$, coding in some appropriate way the labels of each instance of $\X_N$~\cite{MSa2013}.

The $(\x_i,y_i)$ tuples are assumed to be drawn i.i.d. according to a certain probability distribution $\SP(\x,y)$ on $\X\times\Y$~\cite{Vapnik1995,cherkassky2007,MSa2013}.
We also have a \emph{hypothesis space} $\SH$ consisting of functions $h: \X \rightarrow \Y$ and a loss function $L(y, h(\x))$ quantifying the deviation of the predicted value of $h(\x)$ from the true value $y$.
For a given loss function one is able to compute the average loss, or classification risk, $R(h)$, as an expected value of the loss. For absolutely continuous distributions on $\X$ the classification risk (a functional of $h$) is then written as:
\begin{eqnarray}\label{eq:loss}
\nonumber R(h) & \equiv & \SE_{\X\times\Y}[L(\Y,h(\X))] \\
   &=& \sum_{y\in \Y}\int_{\X} \SP(\x,y)L(y,h(\x)) d\x.
\end{eqnarray}
For discrete distributions on $\X$ the classification risk is:
\begin{eqnarray}\label{eq:lossdisc}
\nonumber R(h) & \equiv & \SE_{\X\times\Y}[L(\Y,h(\X))] \\
   &=& \sum_{y\in \Y}\sum_{\x\in\X} \SP(\x,y)L(y,h(\x)).
\end{eqnarray}

Our aim is to derive an hypothesis $h(\x)$ that minimizes $R(h)$. In common practice $\SP(\x,y)$ is unknown to the learner. Therefore, $R(h)$ would be estimated using \meqref{eq:loss} or \meqref{eq:lossdisc} above using an estimate of $\SP(\x,y)$. As an alternative, one could also opt to minimize an empirical estimate of the risk,
\begin{equation}
\label{eq:emploss}
\widehat{R}(h) = \frac{1}{N}\sum_{i=1}^N L(y_i,h(\x_i)).
\end{equation}
Note that if we use an indicator loss function, $L(y,h(\x))=\1_{y\neq h(\x)}$, the risk given by \meqref{eq:loss} or \meqref{eq:lossdisc} corresponds to the probability of error. Finding the function $h$ that minimizes $R(h)$ corresponds then to finding the hypothesis --- also known as decision function --- that minimizes the probability of error. Similarly, $\widehat{R}(h)$ corresponds to an empirical estimate of the probability of error.
Finally, when minimizing \meqref{eq:loss}, $h$ is given according to a parametric form $h(\x,\w)$ with $\w \in A, A\subset \R^n$. Finding the appropriate function means finding its corresponding parameters~\cite{Vapnik1995}.
When clear from the context we will omit $\w$ to define the hypothesis $h(\x)$.

\subsection{Distribution-Based Transfer Learning}\label{sec:unify}
\begin{figure}[!t]
\centering
\includegraphics[width=\columnwidth]{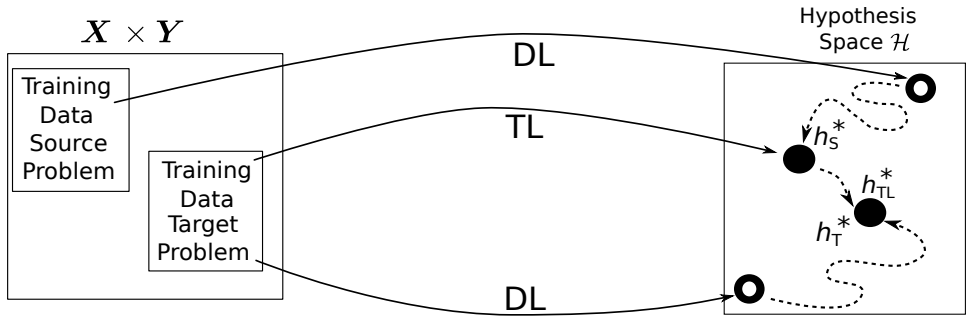}
\caption{A representation of \acf{TL} when we have the same number of classes on both source and target problems. When directly solving the source and target problems (Direct Learning) one usually starts with a random parameterization (open bullets) until attaining the optimal solution, $h^*_S$ or $h^*_T$, respectively.
\ac{TL} uses $h^*_S$ as a starting point to reach $h^*_T$, instead of a random parameterization.}
\label{fig:TL}
\end{figure}

The first requirement is to set the scope of \ac{TL}.
To assess if we can perform \ac{TL} we need to know the data distribution of our source, $\SP_S$, and target, $\SP_T$, problems. We will use subscript $S$ and $T$ to refer to the source and target problems, respectively.

Suppose we have two functions, $h^*_S$ and $h^*_T$, each one solving by \acf{DL} the source and target problems, respectively. While \ac{DL} uses a random initial parameterization, \ac{TL}, by contrast, uses $h^*_S$ as a seed to reach $h^*_T$. This is illustrated in \fref{fig:TL}.

In the work of Shimodaira~\cite{shimodaira2000} a weighted maximum likelihood estimate is devised to handle different data distributions for regression problems.

One can assess this issue theoretically by deriving the target risk w.r.t. the source hypothesis as follows. We start by writing down the risk of the source problem:
\begin{equation}
\SE_S[L(\Y,h_S(\X))] = \sum_{y\in\Y}\int_{\X} \SP_S(\x,y) L(y,h_S(\x)) \d\x,
\end{equation}
where $\SE_S[L(\Y,h_S(\X))]$, $\SE_S$ for short, is the same as $\SE_{\X\times\Y}$ under the distribution $\SP_S(\x, y)$, and we assume that a one-to-one correspondence between source and target spaces exists, denoting the common space by $\X\times\Y$.
The risk minimization process will select an optimal hypothesis, $h^*_S$, with a minimum risk:
\begin{align}
R_S(h^*_S) & \equiv \SE_S[L(\Y,h^*_S(\X))] \nonumber \\
          & = \sum_{y\in\Y}\int_{\X} \SP_S(\x,y) L(y,h^*_S(\x)) \d\x  && \nonumber \\
	 & = \sum_{y\in\Y}\int_{\X} \frac{\SP_S(\x,y)}{\SP_T(\x,y)}\SP_T(\x,y) L(y,h^*_S(\x)) \d\x  && \nonumber  \\
	 & = \SE_T\left [\frac{\SP_S(\X,\Y)}{\SP_T(\X,\Y)} L(\Y,h^*_S(\X)) \right ] \equiv R_T^W(h^*_S),   && \label{eq:TLfinal}
\end{align}
where $W$ stands for weighted.
The equations for discrete distributions are obtained by simple substitutions of the integrals by the appropriate summations.
Assessing the \ac{TL} advantage corresponds to assessing a ``distance'' between the initial solution with risk $h^*_S(\x)$ and the optimal solution one would obtain by \ac{DL} (see \fref{fig:TL}). An appropriate ``distance'' is the deviation of the corresponding risks.
How to assess \ac{TL} has been intensively pursued in psychology. Of most interest is how to measure gains of \ac{TL}. This phenomenon was addressed in~\cite{Perkins1992transfer} where concepts of positive and negative transfer were introduced.
\begin{Def}[Transference]\label{def:transference}
  Transference is a property that allow the measurement of the effects of a \ac{TL} framework (e.g., performance).
  \begin{itemize}[font=\bfseries, style=nextline,leftmargin=0.25cm]
  \item[--] \emph{Positive Transference:} occurs when a \ac{TL} method results on improving the performance on a target problem w.r.t. the best model obtained by \ac{DL} on the target problem;
  \item[--] \emph{Negative Transference:} the opposite of positive transference that is, when a \ac{TL} method results on degrading performance on a target problem.
  \end{itemize}
\end{Def}

\subsection{Categorizations of Classifier Transfer Learning}\label{sec:TLcat}
Based on the joint probability and by the Bayes rule,
\begin{equation}
\label{eq:joint}
\SP(\x,y) = \SP( y | \x ) \SP(\x),
\end{equation}
for the source and target problems, we now generate all TL possibilities.
These possibilities arise from the decision functions and data distribution changes that can occur on the source and target problems. Take the example of \emph{covariate shift}.
If we generate datasets for the source and target marginal distributions, $\SP_S(\x)$ and $\SP_T(\x)$, according to two Gaussian distributions with different means and covariances, and superimpose the same decision function such that a coding $y$ for an instance, $\x \in \R^2$, is assigned according to the rule $d = (x_1-0.5)(x_2-0.5) \text{ with } y = {-1}, d<-1 \text{ and } y={+1} \text{ otherwise}$, we are then able to obtain the source and target datasets shown in \fref{fig:covariateshift}.
\newsavebox{\smlmata}
\savebox{\smlmata}{$\bigl[\begin{smallmatrix}0 & 1 \\ 1 & 0\end{smallmatrix}\bigr]$}
\newsavebox{\smlmatb}
\savebox{\smlmatb}{$\bigl[\begin{smallmatrix}1 & 0.2 \\ 0.8 & 1\end{smallmatrix}\bigr]$}
\begin{figure}[]
\centering
\includegraphics[width=\columnwidth]{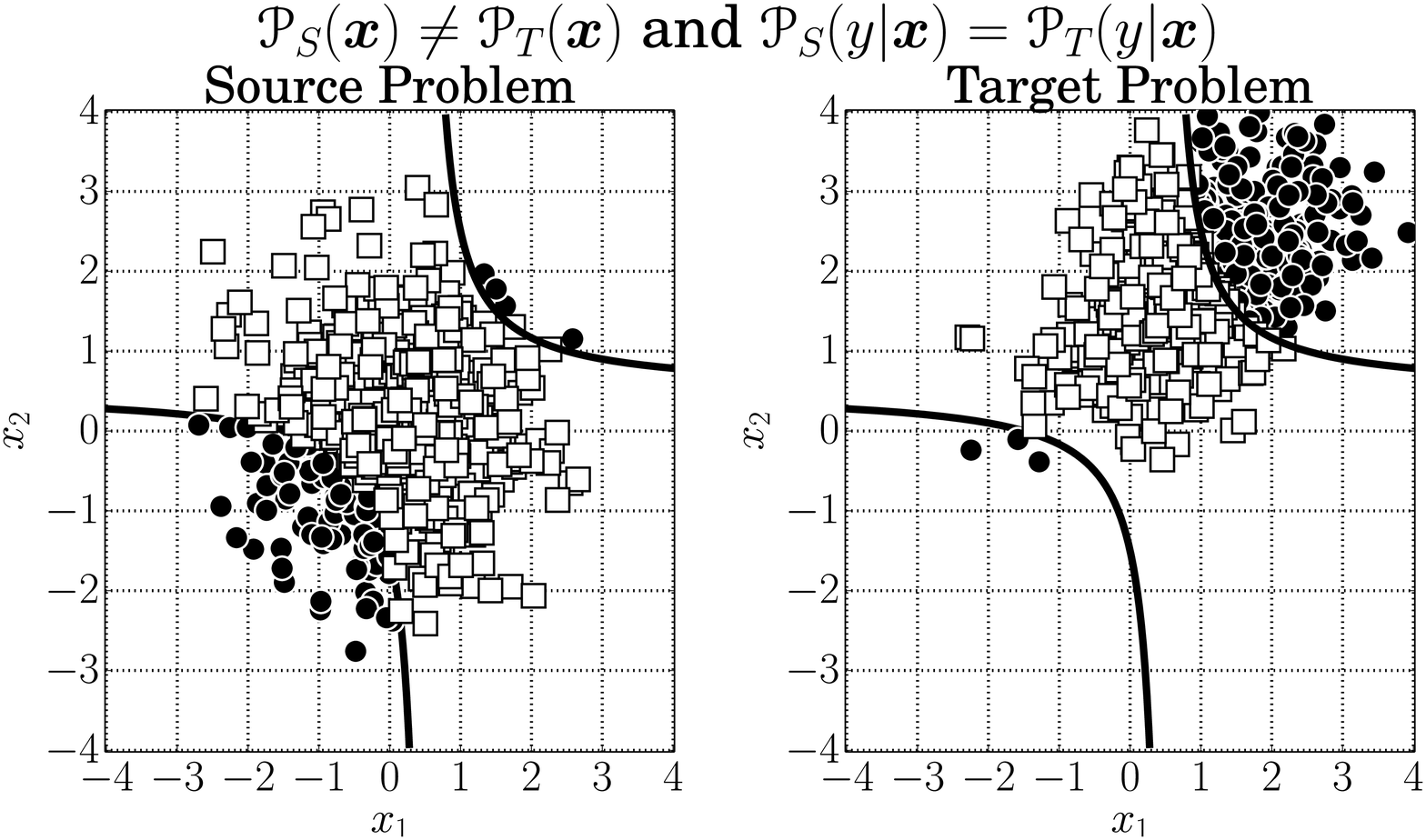}
\caption{A possible scenario for \emph{covariate shift}. Data for the marginal distributions $\SP_S(\x)$ and $\SP_T(\x)$ was generated from two Gaussian distributions with $\mu_S = (0,0)^t$ and $\Sigma_S = \usebox{\smlmata}$ and $\mu_T = (1,2)^t$ and $\Sigma_T = \usebox{\smlmatb}$. The following decision rule was superimposed: $d = (x_1-0.5)(x_2-0.5), y = {-1}, d<-1 \text{ and } y={+1} \text{ otherwise}$.}
\label{fig:covariateshift}
\end{figure}

Inspired by the covariate shift setting, we may then impose conditions on the marginal and posterior distributions (\meqref{eq:joint}), in order to arrive at all possible TL categories presented in \tref{tab:TL}. Note that under a practical perspective it makes sense to use marginal and posterior distributions for TL categorization. Marginal distributions are easy to estimate and histogram inspection may hint on whether or not the posteriors are the same. A practical assessment of the TL category at hand is then achievable.
\rowcolors{1}{gray!25}{white}
\begin{table*}[]
\caption{Multiple possibilities for having \ac{TL}: based on the equality ($ \SP_S(\x,y) = \SP_T(\x,y) $) or not ($ \SP_S(\x,y) \neq \SP_T(\x,y) $) of the source and target problems.}
\label{tab:TL}
\footnotesize
\centering
\begin{tabular}{>{\centering\arraybackslash}p{3.5cm} | >{\centering\arraybackslash}p{4cm}|>{\centering\arraybackslash}p{3cm}}
\rowcolor{gray!50}
\hline\hline
\rowcolor{gray!50}
                                                                    & $ \SP_S(\x,y) = \SP_T(\x,y)$  & $ \SP_S(\x,y) \neq \SP_T(\x,y)$ \\\hline\hline
$\SP_S(\x) = \SP_T(\x)$ and $\SP_S(y|\x) = \SP_T(y|\x)$       & {\bf No \ac{TL}}: everything is the same & {\bf No \ac{TL}}: Impossible\\ \hline

$\SP_S(\x) \neq \SP_T(\x)$ and $\SP_S(y|\x) = \SP_T(y|\x)$    & {\bf No \ac{TL}}: Impossible & {\bf \ac{TL}}: Covariate Shift \\ \hline

$\SP_S(\x) = \SP_T(\x)$ and $\SP_S(y|\x) \neq \SP_T(y|\x)$    & {\bf No \ac{TL}}: Impossible & {\bf \ac{TL}}: Response Shift  \\ \hline

$\SP_S(\x) \neq \SP_T(\x)$ and $\SP_S(y|\x) \neq \SP_T(y|\x)$ & {\bf No \ac{TL}}: Valid but \ac{TL} only for a few particular cases & {\bf \ac{TL}}: Complete Shift\\ \hline
\end{tabular}
\end{table*}
It is now clear that if $\SP_S(\x,y) = \SP_T(\x,y)$ there is no reason to perform \ac{TL}.
The interesting TL settings correspond to $\SP_S(\x,y) \neq \SP_T(\x,y)$ leading to three possible categories:
\begin{description}[font=\bfseries, leftmargin=3cm,style=nextline]
\item[Covariate Shift:] $\SP_S(\x)$ is different from $\SP_T(\x)$ and $\SP_S(y|\x)$ equal to $\SP_T(y|\x)$
\item[Response Shift:]  $\SP_S(\x)$ is equal to $\SP_T(\x)$ and $\SP_S(y|\x)$ different from $\SP_T(y|\x)$
\item[Complete Shift:]  $\SP_S(\x)$ is different from $\SP_T(\x)$ and $\SP_S(y|\x)$ different from $\SP_T(y|\x)$
\end{description}
Different attempts have been made to clarify each of these categories~\cite{Marcu2006,jiang2007instance,scholkopf2009,Pan2010}.
Jiang \etal in~\cite{jiang2007instance}, identify Covariate Shift as Instance Weighting.
In a recent work, Zhang \etal~\cite{Zhang2013} have proposed to categorize \meqref{eq:TLfinal} but based on the class prior, $\SP(y)$, and likelihood, $\SP(\x|y)$, distributions.
Their rationale is not related to the decision surfaces as we propose, thus leading to a different analysis. Moreover, they require further assumptions on the data distributions~\cite{Zhang2013}. For an in-depth survey please refer to~\cite{jiang2008literature}. Finally, and to add more confusion, each of these approaches also encompass different terminologies.
We now proceed to analyze each case.
%

\subsubsection{Covariate Shift}\label{sec:CovariateShift}

For \emph{covariate shift} the following conditions hold:\\ $\SP_S(\x)\neq \SP_T(\x)$ (different marginals),\\ $\SP_S(y|\x)=\SP_T(y|\x)$ (equal posteriors).\\ As a consequence of these two conditions the likelihoods are different ($\SP_S(\x|y)\neq \SP_T(\x|y)$) as well as the joint distributions ($\SP_S(\x,y)\neq \SP_T(\x,y)$).
Note that the equality of the posteriors implies the same optimal decision functions for both the source and the target problems (see e.g.,~\cite{devroye1996}).

Consider now Example \ref{ex:CovariateShift} as a simple illustration of \emph{covariate shift}:
\begin{Example}\label{ex:CovariateShift}
In this example we assume: a univariate $\X\in\{-2,-1,0,1,2\}$; a binary decision problem with $\Y \in \{{-1}, {+1}\}$; and, equal priors for both the source and target problems, $\SP_S({-1}) = \SP_S({+1}) = \SP_T({-1}) = \SP_T({+1}) = 1/2$.

Assume we know the likelihoods of the source problem, say,\\ $\SP_S(x | {-1}) = \{0.4, 0.3, 0.15, 0.1, 0.05\}$ and\\ $\SP_S(x|{+1}) = \{0.05, 0.1, 0.15, 0.3, 0.4\}$.\\ We are then able to trivially compute\\ $\SP_S(x) = \{0.225, 0.2, 0.15, 0.2, 0.225\}$.\\ Note that the symmetry of the likelihoods implies symmetry of the posteriors.
Assume further, that we do not know the likelihoods of the target problem except that they are symmetrical, hinting at a \emph{covariate shift} scenario for \ac{TL}. We assume the marginal distribution to be known:  $\SP_T(x)=\{0.3, 0.19, 0.02, 0.19, 0.3\}$. We see that $\SP_S(x)\neq\SP_T(x)$ as in the \emph{covariate shift} scenario.

We solve the source problem by finding the optimal separating point, by symmetry $x^*_S=0$, such that\\ $\SE [L(\Y,h^*_S(\X))]$ is minimum, with\\ $L(y,h^*_S(\x))=\1_{y\neq h^*_S(\x)}$. The risk is the probability of error:
\begin{align}
R_S(x^*_S)   & = \sum_{y\in\Y}\sum_{x\in\X} \SP_S(x,y) L(y,h^*_S(x)) \nonumber  \\
          & \hspace{-2em} = \frac{1}{2}\sum_{x\in\{{-2},{-1}\}} \SP_S(x|{+1})L({+1}, h^*_S(x)) + \nonumber \\
          &  + \frac{1}{2}\sum_{x\in\{0,1,2\}} \SP_S(x|{-1}) L({-1}, h^*_S(x)) = 0.225. \label{eq:covariateshifteq1}
\end{align}
Let us now work out the target problem. Under the assumption of \emph{covariate shift} the posteriors are equal to those of the source problem, therefore we know that the optimal solution is also $x^*_T=0$. We should then obtain the same value, when using the above equation. In order to check this we need the likelihoods, satisfying the Bayes rule $(\SP(x|y) = \SP(y|x) \SP(x) / \SP(y))$ with the constraint $\sum_x \SP(x|y) = 1$. They are:\\ $\SP_T(x|{-1}) = \{0.5(3), 0.285, 0.02, 0.095, 0.06(6)\}$ and\\ $\SP_T(x|{+1}) = $\mathlist{0.06(6), 0.095, 0.02, 0.285, 0.5(3)}.\\
Algebraically we see that initial assumptions hold and $R^W_T(x^*_S) = R_S(x^*_S) = 0.225$.

We may now question if for this example \ac{TL} was beneficial.
For this we must determine how much \ac{TL} deviated from the target solution if we learn it directly.
Assessing the difference between using $x^*_S$ as a seed to reach $x^*_T$, and \ac{DL}:
$\left | R^W_{T}(x^*_S) - R_T(x_T^*) \right | = 0.053(3),$
and comparing when choosing by chance:\\
$\frac{1}{5} \sum_{x\in\X} \left | R(x) - R_T(x_T^*) \right |  = 0.195,$\\
we see that for this example \ac{TL} rendered a good solution.
\demo
\end{Example}

\emph{Covariate shift}~\cite{Bengio2011,PereiraML2010,Bruzzone2010,shimodaira2000} (inaccurately coined as transductive \ac{TL} in~\cite{Pan2010}) is the most studied category of \ac{TL}~\cite{bendavid2010}.

The conditions for \emph{covariate shift} have also been explored in what is called \ac{DA} (see e.g.,~\cite{Marcu2006,Mansour2009}) with differences on the learning algorithms employed and usage of unlabeled data.
In~\cite{bendavid2010}, Ben-David \etal clarifies this by proposing two more assumptions that must be considered to guarantee successful domain adaptation algorithms.
\emph{Covariate shift} estimation risks have been extended and validated for cross-validation~\cite{sugiyama2007}, and applied to \ac{PoS} tagging (e.g., noun, verb or adjective) or parsing~\cite{Pereira2007,PereiraML2010,Bengio2011}.
In the latter works, \emph{covariate shift} was applied for \ac{PoS} tagging on 100 thousand of Wall Street sentences against the 200 thousand sentences of the biomedical MEDLINE~\cite{Pereira2007,Blitzer2006} or in SPAM dataset~\cite{jiang2007instance}; or, in~\cite{Bengio2011}, where a \ac{DNN} was used for sentiment analysis on the Amazon dataset with two thousand samples on each domain.
In Bickel \etal~\cite{Bickel2009} the \emph{covariate shift} principle is used to identify spam emails, discriminate \ac{ML} from Networking scientific articles and landmine detection (binary classification problems) showing the robustness of this \ac{TL} category.
More recently, Amaral \etal in~\cite{Telmo2014a} employed a \ac{DNN} approach for the recognition of digits using the MNIST and Char74k datasets.
The source problem that was given to the learning model was synthetically generated by rotating the original data, whereas the target problem consisted on the publicly available data. The number of rotations and training data was assessed in their experiments leading to improved results by \ac{TL}.
Covariate Shift (or as referred to as \acf{DA}) has also been explored in an ensemble context with multiple sources~\cite{yao2010boosting} as an improvement of~\cite{YuICML2007}. The former, however, goes beyond the definition for \ac{TL} on classification problems (see Definition \ref{def:TL}). With a different view, Kulis \etal in~\cite{kulis2011you} present an algorithm for finding a common feature representation based on kernel transformations. Although this work fits in a covariate shift scenario, the \ac{TL} problem is reduced on finding a good mapping (linear or non-linear) to represent both source and target problems in the same feature space.
All of the aforementioned methods try to avoid a direct estimation of the densities to guide the learning algorithm. In fact, and as it was mentioned in~\cite{scholkopf2006}, it is excessive to perform such computations just for a weight contribution in each (source and target) problems.

\subsubsection{Response Shift}
\emph{Response shift} is the category of \ac{TL} whose goal is to adapt the existing model to the changes of the \emph{concepts} (classes).\footnote{We have defined that the number of classes between source and target problems would be identical. Extensions for this \ac{TL} category should be considered in the future.}

For \emph{response shift} (also coined as labelling adaptation~\cite{jiang2007instance}) the following conditions hold: $\SP_S(\x) = \SP_T(\x)$ (equal marginals), $\SP_S(y|\x)\neq\SP_T(y|\x)$ (different posteriors). As a consequence of these two conditions the likelihoods are different ($\SP_S(\x|y)\neq \SP_T(\x|y)$) as well as the joint distributions ($\SP_S(\x,y)\neq \SP_T(\x,y)$).

Consider now Example \ref{ex:ResponseShift} as a simple illustration for the \emph{response shift}:
\begin{Example}\label{ex:ResponseShift}
Assume the same source problem as in Example \ref{ex:CovariateShift} with $R_S(x^*_S) = 0.225$ (see \meqref{eq:covariateshifteq1}).

Let us now work out the target problem.
Assume that we do not know the likelihoods of the target problem but, since we are under the response shift scenario, we know that the marginal distribution is:\\ $\SP_S(x) = \SP_T(x)=\{0.225, 0.20, 0.15, 0.20, 0.225\}$.\\
Since the posteriors are different to those of the source problem, therefore we do not know the optimal solution of the target problem. Nevertheless, according to \meqref{eq:TLfinal} we should obtain the same risk value of the source problem.
In order to check this we need the likelihoods, satisfying the Bayes rule with the constraint $\sum_x \SP(x|y) = 1$. They are:\\ $\SP_T(x|{-1}) = \{0.289, 0.289, 0.2329, 0.1404, 0.0487\}$ and \\$\SP_T(x|{+1}) = $\mathlist{0.1610, 0.111, 0.0671, 0.2596, 0.4013}.\\ Thus, $x_T^* = 1$. Algebraically we see that initial assumptions hold and $R^W_T(x^*_S) = R_S(x^*_S) = 0.225$.

Once again, we may now question if for this example \ac{TL} was beneficial.
For this we must determine how much \ac{TL} deviated from the target solution if we learn it directly. Assessing the difference between using $x^*_S$ as a seed to reach $x^*_T$, and \ac{DL}:
$\left | R^W_{T}(x^*_S) - R_T(x_T^*) \right | = 0.039,$
and comparing when choosing by chance:\\
$\frac{1}{5}\sum_{x\in\X}\left | R(x) - R_T(x_T^*) \right |  = 0.15,$\\
we see that for this example \ac{TL} rendered a good solution.
\demo
\end{Example}

To the best of our knowledge, there are few works that analyze only the \emph{response shift} scenario.
Jiang \etal in~\cite{jiang2007instance} explore the response shift scenario (with the instance weighting approach) to study the labeling bias on the target domain.
In~\cite{scholkopf2009} there was an attempt to this approach where authors introduced a new \ac{SVM} formulation so they could handle different predictor functions.
Their approach was tested on a mRNA problem for the detection of splice sites with 100 thousand examples per problem~\cite{scholkopf2009}.
Amaral \etal in~\cite{Telmo2014b} explored this category of \ac{TL} by assessing the performance of \acp{DNN}.
Using the same data, they do \ac{TL} from digits recognition to a problem of odd vs. even discrimination.
Amaral \etal in~\cite{Telmo2014b} also evaluated \ac{TL} performance from a shape recognition (e.g., rectangles, circles and triangles) to a corner vs. round object recognition target problem, again, using the same data, as in the framework of \emph{response shift}.
In both experiments they analyzed \ac{TL} performance when transferring part of the architecture of the \ac{DNN} as well as when different sample sizes are used for training.

\subsubsection{Complete Shift}
For \emph{complete shift} the following conditions hold:\\ $\SP_S(\x)\neq \SP_T(\x)$ (different marginals),\\ $\SP_S(y|\x) \neq \SP_T(y|\x)$ (different posteriors).\\
Example \ref{ex:CompleteShift} illustrates this category:

\begin{Example}\label{ex:CompleteShift}
Assume the same source problem as in Examples \ref{ex:CovariateShift} and \ref{ex:ResponseShift} with $R_S(x^*_S) = 0.225$.
Let us now work out the target problem.
Assume that we do know the likelihoods of the target problem such that\\ $\SP_T(\x | {-1}) = \{0.6, 0.1, 0.1, 0.1, 0.1\}$ and\\ $\SP_T(\x|{+1}) = \{0.01, 0.2, 0.2, 0.2, 0.39\}$.\\ In the same manner, we are able to trivially compute \\$\SP_T(\x) = \{ 0.45, 0.275, 0.15, 0.07, 0.05\}$.\\
By applying Bayes rule we obtain:\\ $\SP_S({-1} |\x ) = $\mathlist{0.8(8), 0.75, 0.5, 0.25, 0.1(1)} and\\ $\SP_S({+1} |\x ) = $\mathlist{ 0.1(1), 0.25, 0.5, 0.75, 0.8(8)}\\  for the source problem;\\ $\SP_T({-1} |\x )$ = \mathlist{0.9836, 0.3(3), 0.3(3), 0.3(3), 0.2041} and \\$\SP_T({+1} |\x ) = $\mathlist{ 0.0164, 0.6(6), 0.6(6), 0.6(6), 0.7959}\\ for the target problem.
We see that \\$\SP_S ( \x ) \neq \SP_T ( \x )$ and $\SP_S ( y | \x ) \neq \SP_T ( y | \x )$ \\as in the complete shift scenario.
According to \meqref{eq:TLfinal}, we should then obtain the same risk values of the above equation.
Thus, $x_T^* = {-1}$. Algebraically we see that initial assumptions hold and $R^W_T(x^*_S) = R_S(x^*_S) = 0.225$.

Let us analyze again if for this example \ac{TL} was beneficial.
Assessing the difference between using $x^*_S$ as a seed to reach $x^*_T$, and \ac{DL}:
$\left | R^W_{T}(x^*_S) - R_T(x_T^*) \right | = 0.02,$
and comparing when choosing by chance:\\
$\frac{1}{5}\sum_{x\in\X}\left | R(x) - R_T(x_T^*) \right |  = 0.159,$\\
we see that for this example \ac{TL} rendered a good solution.
\demo
\end{Example}

 \emph{Complete shift}is the most comprehensive setting of \ac{TL}. This category of \ac{TL} has also been coined as Never-Ending Learning~\cite{mitchell2015never} or Curriculum Learning~\cite{bengio2009curriculum}.
In~\cite{caputo2014} a weighted scoring model is applied to be adaptable for different source and target problems, while being capable at the same time to learn new classes. The work of Kulis~\cite{kulis2011you} also goes in line with Caputos' work.
In a way, these works~\cite{caputo2014,kulis2011you} revisit the work conducted by Thrun in~\cite{Thrun96a} which has been coined as inductive transfer~\cite{torrey2009transfer,niculescu2007inductive}.
Interesting enough, and to the best of our knowledge,~\cite{caputo2014} is the first work that tries to \emph{unify} concepts that exist in the literature from \ac{TL} although without a theoretical framework. Other approaches exist that assume different distribution and predictive functions~\cite{Tommasi_PAMI2013}.
Nevertheless, these approaches consider that the number of classes can increase over time which goes beyond the scope of this paper.
In~\cite{Chetak2014a,Chetak2014b} \acp{SDA} and \acp{CNN} were employed for \ac{TL} under the \emph{complete shift} assumption.
In both works the MNIST and MADbase datasets were used.
In particular, it was empirically shown that \ac{TL} attained better results if the source problems corresponded to harder problems than simpler ones~\cite{Chetak2014a}.
More recently, in~\cite{habrard2015new} authors present an ensemble algorithm for this specific categorization of \ac{TL}.

\section{Transfer Learning: Open Issues}\label{sec:issues}

Having stressed the importance to unify some of the concepts that have been presented in the literature about TL, we will now focus on other important issues that should be considered in the near time soon. Although being conscious that problems like the difficulty on analyzing huge amount of available data that is continuously increasing or the inexistence of public competitions to benchmark new and available methods of TL are important, we think that there are two issues on transfer learning that one must address: how to measure knowledge gains when doing TL and; how differences among source and target datasets may impact the learning rates.

\subsection{Measuring Knowledge Gains}

In the previous section, we mentioned that measures proposed in~\cite{Bengio2011} were used to analyze and quantify TL gains.
Although clarifying some interpretation issues when analyzing performance results specially when dealing with different multiple source domains, it is not clear how these measures can be used in other TL methods besides covariate shift, specially when class sets are different between the studied problems.
Approaches like mean squared error (MSE) or some statistical inspired coefficients may provide further information such as class agreement. Even though the aspect of MSE not being bounded could be seen as a drawback, it provides rich knowledge on the behavior of TL algorithms.
Moreover, measures as those employed in~\cite{Bengio2011} can lead to non precise results if a perfect baseline model is obtained (e.g., $\remp{T} = 0$ in \meqref{eq:risktransferone} or \meqref{eq:risktransfertwo}).

As mentioned before, one goal of TL is to obtain a model learned on the source domain that can attain similar or improved performance rates on the target domain than if it was learned solely on the target domain. If our empirical risks for our source and target problems are formally given by:
\begin{eqnarray}
  \remp{S} &=& \frac{1}{N_S} \sum_{i=1}^{N_S} L(y_i, h_S (\x_i ) ) \\
 \remp{T} &=& \frac{1}{N_T} \sum_{i=1}^{N_T} L(y_i, h_T (\x_i ) )
\end{eqnarray}
respectively, then the empirical risk for TL is given by:
\begin{equation}
\label{eq:risktransfer}
\remp{TL}  = \remp{T|S} - \remp{T}\\
\end{equation}
given that
\begin{equation}
     \remp{T|S} = 1/N_T sum^{N_T} L(y_i,h*_S(x_i))
\end{equation}
is the empirical risk of our target model learned based on the source model.

To the best of our knowledge,~\cite{Bengio2011} were the first to analyze the concept of transfer error. When dealing with multiple domains becomes difficult to understand the overall result by averaging all results using \meqref{eq:risktransfer} for each source domains. Two measures were thus proposed in [24] to assess the ratio of occurred transference: transfer ratio and in-domain ratio. The former is the straightforward average of the transference ratios among all domains:
\begin{equation}
\label{eq:risktransferone}
\remp{TL} = \frac{1}{SD} \sum_{d=1}^{SD} \frac{\remp{T|S}}{\remp{T}}
\end{equation}

\begin{Def}[Representation function]\label{def:representation}
A representation function $\Phi$ is a function which maps instances to features $\Phi: \SX \rightarrow \SZ$, where $\SZ \in \R^n$.
\end{Def}

Due to the usability of deep networks and their capability to transfer raw features to abstract representations,~\cite{Bengio2011} also assessed the performance on using this information given by
\begin{equation}
\label{eq:risktransfertwo}
\remp{TL} = \frac{1}{SD+TD} \sum_{d=1}^{SD+TD} \frac{\remp{\Phi(T)|\Phi(T)}}{\remp{T}}
\end{equation}
where $\Phi$ is an intermediate representation for the instances, given by~\cite{Pereira2007,PereiraML2010}.

\subsection{Dissimilar Datasets: How difficult is to do Transfer Learning?}
Conventional works categorize different TL methodologies based only on the joint probability ($\SP(\x,y)$) of the source and target problem (e.g.,~\cite{Bruzzone2010}). Despite being too strict (source and target data are defined jointly with the classes) this concept thrived in the literature mostly without major justification.
This definition may be important when data distribution does not affect the problem, such as \emph{dataset shift} referred in~\cite[Chapter 1]{Quionero2009}. On the opposite, on transfer learning our data distribution changes significantly between source and target problems. Recall the examples given in the beginning of this manuscript of performing recommendations of consumer items for the Amazon website and to the IMDB, or the text classification for the Wikipedia website and the Reuters website.
Another aspect that differentiates this work is that we focus on the problem and not on which method was used to solve the TL problem (e.g., random projections, SVMs or deep architectures~\cite{Pereira2007amazon,Bengio2011,Tommasi_PAMI2013}).

An obvious question that can be raised, is how can we measure different domains, therefore, distributions? Much work has been conducted on this matter. Traditionally, Kullback-Leibler (KL) divergence has been used to estimate such differences~\cite{LinTIT1992,YuICML2007}. Given two probability functions $p(\x)$ and $q(\x)$, KL divergence is defined as:

\begin{equation}
D_{KL} (p || q) = \sum_x p(\x) \log \frac{p(\x)}{q(\x)}.
\label{eq:KL}
\end{equation}
Besides the theoretical and practical limitations of this measure (undefined when $q(\x) = 0$) and having no upper bound, one drawback of this measure is that it cannot be defined as a distance since it does not obey the symmetry and triangular inequality properties. An alternative to this is the well known Jensen-Shannon (JS) divergence~\cite{LinTIT1992} given by:

\begin{equation}
D_{JS} (p||q) = \alpha D_{KL} (p||r) + \beta D_{KL} (q||r),
\label{eq:JS}
\end{equation}
with $r = \alpha p + \beta q$, where $D_{KL}$ is the Kullback-Leibler divergence defined in \meqref{eq:KL}. When $\alpha=\beta=1/2$ in \meqref{eq:JS} we are dealing with the \emph{specific} Jensen-Shannon divergence and $D_{JS}$ is lower- and upper-bounded by $0$ and $1$, respectively, when using logarithm base 2~\cite{LinTIT1992}. This means that when $D_{JS} (p||q) = 0$ we can consider that $p$ and $q$ are identical and when $D_{JS} (p||q) = 1$, the distributions are different.

In the TL literature these concepts are not novel, but its usage to discriminate different TL settings are unknown to us, leading to the second contribution of this work.
JS divergence has been mostly used as a tool to validate some assumptions over the tasks or as a \emph{quantity} of the difficulty of learning algorithms~\cite{Bruzzone2010}.

It is important to state that according to~\cite{YuICML2007} the quality of the results is related with KL-divergence measured on the datasets with different domains. In a straightforward reading it may seem that for different pairs source/target problems it may be infeasible to perform TL. Or, TL models need to be more robust for more heterogeneous problems. Or, it can even mean that features of these domains are not representative.
Based on our review, it was not possible to identify works that try to make this analysis or at least to perform an attempt on that.
Although these intuitive ideas have empirically presented a relation between domain divergence and TL algorithms performance, a theoretical reason for these behaviors is still unknown~\cite{Bruzzone2010,Chetak2014a}.

-------------------------------------------------------------------------------------------------------------------------
\section{Conclusions}\label{sec:five}

A study for classification problems with \ac{TL} was presented in this paper. We conducted a review of classical and state-of-the-art work on \ac{TL} by enumerating key aspects of each one. Contrary to the most recent works that progressively conveys Domain Adaptation as a general view of \ac{TL}, in this paper we  presented why the three following categories of \ac{TL} should be considered independently: \emph{covariate shift}, \emph{response shift} and \emph{complete shift}. Each \ac{TL} category was theoretically presented, discussed and illustrated with examples. Beyond the theoretical examples shown we have also outlined the most recent works that fit in each category. The main contribution of the paper is the clarification of \ac{TL} for classification using its principles and concepts which we hope can lead to a clearer understanding of the subject and facilitate the work in this important area.

\section{Compliance with Ethical Standards}

\textbf{Funding}: This work was financed by FEDER funds through the Programa Operacional
Factores de Competitividade COMPETE and by Portuguese funds through
FCT Funda\c{c}\~{a}o para a Ci\^{e}ncia e a Tecnologia in the framework of the project
PTDC/EIA-EIA/119004/2010.\\
\textbf{Conflict of Interest}: Authors declare that they have no conflict of interest.\\
\textbf{Ethical approval}: This article does not contain any studies with human participants or animals performed by any of the authors.

\end{document}